# PIAE: Physics-Informed Continuous Autoencoding for Compact 3D Human Motion

Osama Alshammari[1,2*] Omar Mohammed[1,2*] Kerven Durdymyradov[1] Naeemullah Khan[1,2]

[1]King Abdullah University of Science and Technology (KAUST), Thuwal, Saudi Arabia
[2]Department of Computer Science, King Khalid University (KKU), Abha, Saudi Arabia

osama.alshammari.ai@gmail.com omarhamadevl@gmail.com kerventkm@gmail.com
naeemullah.khan@kaust.edu.sa

**Abstract**

Temporal compression reduces the sequence length that motion generators must model, but can damage articulated detail and global trajectory. These errors have unequal consequences: root-velocity and heading errors propagate through trajectory recovery and affect the placement of the entire body. We propose PIAE, a physics-informed continuous autoencoder that allocates a fixed token budget between explicitly retained kinematic signals and complementary learned features. At 8 × temporal compression, each 256-dimensional token contains 224 learned values and a 32-dimensional physical carrier. The complete token conditions neural decoding, while a parallel recovery path constrains selected outputs. On the official HumanML3D benchmark, PIAE achieves 8.936 mm global MPJPE, an 82.0% reduction relative to a same-width learned-only model. The benefit extends beyond trajectory restoration: facing-local neural-body error before analytical correction decreases by 76.6% in a frozen-checkpoint validation diagnostic, and supplying the same true trajectory to both models preserves the body-quality advantage. Targeted root-only and explicit-root hybrid controls establish strong trajectory-aware comparisons; alternative carriers reveal that torso references favor position accuracy, whereas stance anchors favor support quality. With the original 8× AEs frozen, matched text-to-motion generators predict every carrier value and improve FID from 0.506 to 0.163. Frozen reconstruction also transfers to AIST++ without adaptation. The compressed carrier interface reduces measured generation time by 27.8% relative to its 1 × counterpart. Together, these findings demonstrate that selective preservation can use a fixed representation budget more effectively, improving neural articulated reconstruction and downstream generation in compact human motion.

Code: `https://github.com/EviLooo/PIAE`

## 1 Introduction

A reconstructed character can retain recognizable walking poses yet move along the wrong path. Motion errors have unequal consequences: a local joint-coordinate error perturbs articulation, whereas a persistent root-velocity or heading error can displace or rotate every subsequent pose. A compact motion autoencoder must preserve body detail and global movement despite this unequal sensitivity. Compression matters beyond storage: reducing the temporal sequence also reduces the number of positions and attention interactions a downstream generator must model. A useful representation must deliver those savings while retaining information that is important for reconstruction and generation.

We investigate a simple allocation question: *under the same total representation budget, can explicitly preserving selected kinematic information improve the remaining neural reconstruction and the usefulness of the representation for generation?* Rather than asking the learned bottleneck to approximate every quantity, we reserve capacity for signals that define the trajectory or provide shared spatial and support references, and learn complementary motion features. The explicit values replace learned dimensions. Any benefit must therefore justify their use of scarce capacity, rather than follow from increasing the representation size.

Learned motion latents support text-conditioned synthesis in MLD and MotionGPT[4, 7], while ParTY and PRISM study structured tokenization and latent organization [ 10, 11]. ARDY combines explicit global root information with a learned body code and root-conditioned reconstruction for interactive generation [12]. PIAE studies the complementary representation-design question of how explicit allocation and compatible recovery affect accuracy and downstream utility at a fixed total width. An adapted ARDY-style autoencoder provides a trajectory-aware comparison within the same HumanML3D training and token-budget contract.

*These authors contributed equally to this work.

We introduce PIAE, a physics-informed continuous autoencoder with an explicit in-token *carrier*. At 8× temporal compression, an eight-frame block becomes one 256-dimensional token: 224 learned values and 32 carrier values retaining planar-root and yaw velocities, height summaries, and soft support cues. The full token is mixed and processed by the neural decoder, so the carrier can inform articulated reconstruction as well as trajectory recovery. A parallel path reads the carrier to restore velocities and constrain selected height/contact outputs. Both paths participate during AE training, before the reconstruction losses are calculated. During text-to-motion generation, the generator predicts the entire token from text and noise, including all carrier values.

Three findings connect the design to its practical value. First, the improvement reaches the neural body pathway: at 8×, facing-local body error falls from 40.011 to 9.352 mm before analytical correction. Giving both frozen models the same true trajectory still leaves a large body-quality gap. Second, the representation is useful when its explicit information must be generated rather than extracted from an input: a matched 8× generator improves FID from 0.506 to 0.163. Third, the original frozen AE retains a reconstruction advantage on AIST++ without fine-tuning or normalization refitting. Together, these tests establish neural reconstruction, downstream-task use, and cross-dataset transfer as distinct benefits of the implemented representation.

The controlled studies also explain how to interpret those benefits. Root/yaw retention provides a strong foundation, and the full PIAE design improves on both the root/yaw-only and adapted explicit-root controls. In a separate same-width study, torso references give the best positional accuracy, while stance anchors best improve the measured support behavior. Exact-height replacement is less effective than the original height/support design. These outcomes show why preservation must be considered together with the learned capacity it replaces and the recovery rule it supports. The central takeaway is a division of representational roles: selected explicit information provides reliable references, while the neural pathway reconstructs complementary detail.

Our contributions are:

- **Physically-informed latent tokens (the carrier).** We develop a fixed-width continuous representation with complementary explicit and learned components. The same carrier informs neural decoding and structured recovery; at 8×, its 32 values count within the 256-value token budget.
- **Improved neural reconstruction beyond trajectory recovery.** Same-budget ablations and a frozen validation diagnostic establish a 76.6% reduction in neural facing-local body error at 8 × before correction. A common-trajectory comparison confirms that post-hoc path repair alone does not reproduce the carrier model's body quality.
- **Outcome-dependent preservation under a shared budget.** Root/yaw-only, adapted explicit-root, and alternative-carrier studies examine the allocation-and-recovery design. Torso references improve positional accuracy most, whereas stance anchors best improve support metrics, identifying distinct uses for explicitly retained information.
- **Generative utility, transfer, and measured efficiency.** With the original AEs frozen, predicting the complete 8 × carrier token improves matched text-to-motion FID by 67.7%. Frozen AIST++ reconstruction demonstrates cross-dataset transfer, while the compressed carrier interface reduces measured generation time by 27.8% relative to its 1× counterpart.

# 2 Related Work

**Learned motion representations.** Continuous motion representations include Transformer variational autoencoders and the VAE used for latent diffusion in MLD [3, 4]. Discrete systems apply vector quantization [5]: T2M-GPT and MotionGPT model tokens autoregressively, while MoMask and MMM use masked modeling [6, 7, 8, 9]. These methods establish useful interfaces for generation but make different choices about temporal rate, code organization, and learning objectives. PIAE instead studies deterministic continuous autoencoding with a specified division between learned and explicitly retained values.

**Explicit-root and structured hybrids.** ARDY patchifies global root position and heading, combines them with a learned body code, and conditions body reconstruction on root information for interactive generation [12]. PIAE uses per-frame planar/yaw velocities plus height and support summaries within a 256-value, eight-frame token, combining neural conditioning with compatible analytical recovery. The resulting design supports exact canonical planar-trajectory recovery and provides a setting for studying how the allocation affects neural body quality, generated tokens, and frozen transfer. We evaluate an adapted ARDY-style root–body autoencoder under the common budget; its implementation and comparison scope are specified in Section 4.2.

ParTY studies part guidance and temporal-aware tokenization, while PRISM decomposes its latent by joint structure [10, 11]. Their factorizations differ from PIAE's explicit/learned split, but share the motivation to organize a representation according to the structure of motion. Our experiments distinguish common-protocol reconstruction reruns from published reference values, and count all learned and carrier values in each continuous token.

**Kinematic constraints and recovery.** HumanML3D exposes local pose, root-motion, and contact-related information [1]. PIAE uses their physical meaning to select token contents and reconstruction rules. Here, *physics-informed* refers to kinematic and geometric structure and reconstruction objectives, not rigid-body simulation or a complete dynamical model. The frozen body-path diagnostic specifically tests whether these explicit signals are associated with better neural articulated output in addition to structural path retention.

**Downstream generation and transfer.** We adapt MotionHiFlow's hierarchical temporal generator and use its official evaluator [13], keeping the generative architecture fixed across the four original AE interfaces to test their downstream utility. AIST++ provides music-aligned SMPL dance motions [14]; its official pose-test assets supply a separate frozen reconstruction test with unchanged HumanML3D normalization.

# 3 Method

PIAE assigns different roles within one fixed token. Root/yaw velocities are retained because integration can propagate their errors through the global motion; a learned code supplies articulated detail, conditioned on those values and partial height/support information. Both parts occupy the same budget. The analytical path guarantees recovery only of the quantities it explicitly stores and constrains, not of the complete motion. At generation time the token must instead be predicted, so input-trajectory preservation is not a ground-truth trajectory guarantee for text-to-motion.

## 3.1 Native motion representation

Let $P \in \mathbb{R}^{T\times 22\times 3}$ denote a motion sequence sampled at $f = 20$ Hz. We transform the HumanML3D joints to a canonical, $z$-up coordinate frame with initial planar-root position and heading set to zero. This removes an initial coordinate offset, not the subsequent trajectory. For each frame, we construct

$$x_t = [p_t^{\text{local}}, v_t^{xy}, \omega_t, h_t] \in \mathbb{R}^{66+2+1+1}, \tag{1}$$

where $p_t^{\text{local}}$ contains yaw-local, root-relative coordinates for 22 joints, $v_t^{xy}$ is planar-root velocity, $\omega_t$ is yaw angular velocity, and $h_t$ is root height. Means and standard deviations computed from training data normalize these 70 channels. Contact targets supervise the output but are not additional encoder inputs.

## 3.2 Masked temporal autoencoder

Three encoder stages each apply two masked Transformer blocks and stride-two temporal downsampling. The resulting temporal length is $L = T/8$. Each stage uses hidden width 256 and eight attention heads. Positional encoding distinguishes time indices, while masks exclude padded frames from attention and feature processing. The learned output is projected to 224 dimensions per token. Training lengths are multiples of eight between 40 and 192 frames; batch padding consists of fully padded eight-frame blocks.

The decoder projects the complete 256-dimensional token and reconstructs the temporal resolution through three stride-two upsampling stages with masked Transformer processing. Separate heads predict

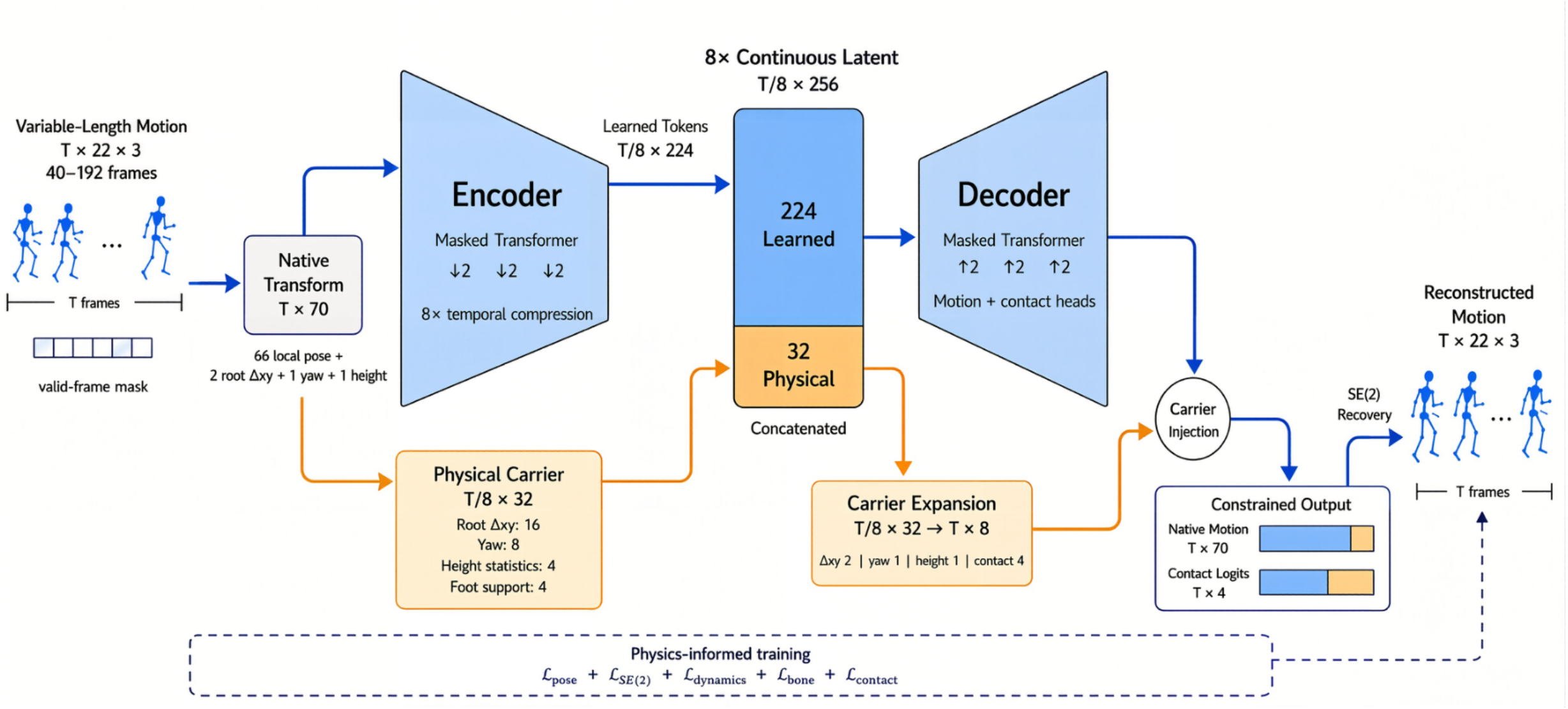

**Figure 1:** PIAE architecture. Every eight frames produce one continuous token with 224 learned and 32 carrier values. The complete token conditions the decoder, while carrier expansion directly constrains selected outputs. In this schematic, "root $\Delta xy$" and "yaw" denote planar velocity (m/s) and yaw angular velocity (rad/s), respectively. Height statistics and support cues are partial summaries, not lossless full-frame height or contact storage. $T$ denotes a valid length divisible by eight; padding is masked.

**Table 1:** Carrier allocation within each 256-dimensional token. All 32 values count toward the latent budget.

| Signal | Values | Use in reconstruction |
|---|---|---|
| Planar-root velocities | 16 | Eight 2D velocities, recovered directly |
| Yaw angular velocities | 8 | Eight angular velocities, recovered directly |
| Root height: start, end, mean | 3 | Endpoint- and mean-preserving height base |
| Mean soft foot support | 4 | One prior per foot joint over the block |
| Root-height standard deviation | 1 | Additional neural-decoder conditioning |
| Total | 32 | Combined with 224 learned values |

70 native-motion channels and four contact logits. Figure1 shows the learned and analytical reconstruction paths. The model contains approximately 22.59 million parameters.

## 3.3 Explicit physical carrier

For eight-frame block $B_k$, a deterministic function $C$ computes a carrier from the unnormalized motion:

$$z_k = [\ell_k; c_k], \qquad \ell_k \in \mathbb{R}^{224}, \quad c_k = C(x_{B_k}) \in \mathbb{R}^{32}. \tag{2}$$

Table 1 specifies its contents. The planar and yaw velocities retain every frame in the block. Height and support information instead retain selected statistics. Fixed physical scales condition the carrier numerically before concatenation; these are reversed when expanding it.

Soft support confidence is computed from the input feet's height and speed using smooth threshold functions, then averaged within the block. Thus it provides motion-derived support information to the decoder without external contact-label inputs.

## 3.4 Dual-path reconstruction and trajectory recovery

The neural pathway consumes $[\ell_k; c_k]$ jointly. The analytical pathway expands $c_k$ into frame-level velocities, a height base, and four support priors. It overwrites the predicted planar and angular velocities with the stored values. Root height combines a base that preserves the encoded endpoints and mean with a learned residual projected to have zero endpoints and zero block mean. The carrier's height standard deviation conditions the neural pathway but is not enforced analytically. Support probabilities provide an additive logit prior to the contact head.

To recover the global trajectory, let $r_t \in \mathbb{R}^2$ and $\theta_t$ be planar-root position and yaw, and let $R(\theta)$ be a planar rotation. In our convention,

$$\theta_t = \theta_{t-1} + \omega_t / f, \tag{3}$$

$$r_t = r_{t-1} + R(\theta_{t-1}) v_t^{xy} / f. \tag{4}$$

The first-frame velocities are zero, with $r_0 = 0$ and $\theta_0 = 0$. Global joints follow by rotating local joints and adding the root position and height.

If the recovered $v_t^{xy}$ and $\omega_t$ equal their encoded input values, Equation 4 reproduces the canonical planar trajectory by induction, subject to floating-point error. PIAE therefore preserves the planar path structurally, while learning articulated reconstruction and within-block height detail. All required carrier information is contained in the encoded token.

### 3.5 Selective preservation under a fixed budget

More generally, a continuous token can be written as

$$z_k = [E_\theta(x)_k; C_{\mathcal{S}}(x)_k] \in \mathbb{R}^{D-d_c} \times \mathbb{R}^{d_c}, \tag{5}$$

where S selects the signals to retain and $d_c$ specifies their storage allocation. The selection is an architectural design choice; $d_c$ is a capacity hyperparameter. At fixed total width $D$, a larger carrier leaves fewer dimensions for learned information.

Our primary choice $D = 256$, $d_c = 32$ prioritizes trajectory-defining velocities, supplemented by vertical and support summaries. This reflects the different consequences of error: velocity errors accumulate through recovery, whereas many local-pose details can be reconstructed directly. The design does not require encoding every physical property. More generally, spatial reference signals may have high reconstruction impact without being temporally integrated. We test this distinction with torso and support anchors in Section 5.4. Their extraction and compatible correction rules are specified in Appendix A.3; every additional carrier dimension displaces a learned dimension rather than increasing total token width.

The experiments specify S and $d_c$ as architectural choices. Each preserved signal has a defined extractor, numerical scale, and compatible recovery rule, and every stored value counts toward $D$. This formulation links allocation to use: the complete representation should support accurate neural body reconstruction, and a downstream generator must predict both explicit and learned values. We evaluate these properties through the body-path diagnostic, alternative-carrier study, and matched generated-token experiments.

### 3.6 Physics-informed reconstruction objective

Training combines native-feature reconstruction, global-joint reconstruction, temporal matching, and skeletal/contact constraints. Let $\text{MSE}_M$ denote a component-wise squared-error mean over valid masked entries. The principal joint and temporal terms are

$$\mathcal{L}_{\text{global}} = \text{MSE}_M(\hat{P}, P), \tag{6}$$

$$\mathcal{L}_{\text{vel}} = \text{MSE}_{M'}(f\Delta\hat{P}, f\Delta P), \tag{7}$$

$$\mathcal{L}_{\text{acc}} = \text{MSE}_{M''}(f^2\Delta^2\hat{P}, f^2\Delta^2 P). \tag{8}$$

The interval masks require all contributing frames to be valid. These terms match target dynamics rather than penalizing acceleration simply for being large. Bone-length error uses an absolute difference. Contact supervision includes binary cross-entropy, target-contact-gated squared planar foot speed, and contact-gated foot-position matching. A squared below-ground penalty discourages penetration. Appendix A provides all weights and implementation details. Auxiliary losses on the directly preserved velocities and terminal planar motion are numerically negligible under the carrier constraint.

### 3.7 Compression accounting

PIAE reduces the number of temporal positions by $8\times$: a 192-frame motion becomes 24 tokens. It stores $256/8 = 32$ continuous scalars per frame on average. The FP32 latent payload is $24 \cdot 256 \cdot 4 = 24{,}576$ bytes (24 KiB) per 192-frame clip, versus $192 \cdot 256 \cdot 4 = 196{,}608$ bytes (192 KiB) for the 1× reference. Relative to the 70-channel native representation, the scalar-count ratio is

$$\frac{8 \cdot 70}{256} = 2.1875\times . \tag{9}$$

Relative to raw 66-channel joint coordinates, the corresponding ratio is 20625×. The 8× storage reduction above is specifically relative to our same-width 1 × latent; the native-to-latent reduction is 2 .1875× at equal FP32 precision. These are representation-dependent scalar counts, and a continuous token cannot be equated in bitrate with a discrete codebook index without specifying precision and coding. Every carrier value is included in the reported count.

For a full self-attention layer operating on the motion latents, the pairwise score matrix has $L^2$ entries. Holding token width fixed while changing the temporal rate from 1× to 8× changes this count from $T^2$ to $(T/8)^2 = T^2/64$. For example, a 64-frame motion changes from 4 ,096 possible motion-token pairs to 64. This concerns the attention matrix of a comparable layer on the compressed sequence, not the FLOPs or memory of the complete system: PIAE's encoder and decoder still process higher-resolution stages, and other generator operations have different costs. We therefore measure end-to-end inference separately.

# 4 Experimental Protocol

## 4.1 HumanML3D data and evaluation population

We use HumanML3D's official splits and 22-joint motion at 20 FPS [ 1]. The dataset draws on motion-capture sources including AMASS [2]. Training uses only the official training split, and checkpoint selection uses validation MPJPE. The reported autoencoders are exponential-moving-average checkpoints trained for up to 300 epochs.

The official test split has 4,384 identifiers. The shared reconstruction contract requires at least 40 frames, excluding 174 shorter clips and leaving 4,210 eligible clips. Every model receives the same longest centered crop divisible by eight, capped at 192 frames. All 4,210 clips are evaluated with zero skips and without padding short motions into eligibility.

## 4.2 Controlled temporal-rate and carrier variants

We train four models from scratch under the same optimization and split contract. The 8× models produce one 256-dimensional token per eight input frames, or 32 latent scalars per motion frame. With the carrier enabled, that token contains 224 learned and 32 physical dimensions; without it, all 256 dimensions are learned. The 1× references produce one 256-dimensional token per frame. Their carrier split is 248 learned plus 8 physical dimensions, because a single frame needs two planar-root velocities, one yaw velocity, one height value, and four support values. Thus the carrier/no-carrier comparison at a fixed rate has equal width and storage, whereas 1× versus 8× is deliberately a compression comparison: the 1× representation stores eight times as many latent scalars.

We additionally train two targeted 8× autoencoders from scratch. The *root/yaw-only* variant retains all eight planar-root and yaw velocities (24 physical plus 232 learned values per token), leaving height and contact to the learned decoder. The *ARDY-style* variant retains each frame's canonical global root $xyz$ and heading sine/cosine (40 explicit plus 216 learned values), with a root-conditioned body decoder. This is our adaptation of ARDY's root–body representation to the HumanML3D pipeline, not its released tokenizer or complete generator. All three 8 × explicit variants have the same total width, rate, dataset, crop, normalization, optimization schedule, seed, and loss weights. Both new runs complete 300 epochs; EMA checkpoints are chosen by official validation MPJPE before the test evaluation. The same 4,210 test clips, coordinate frame, and reconstruction metrics are then used with zero skips. These controls compare complete allocation-and-reconstruction designs under a common budget.

## 4.3 Frozen neural-body diagnostic

Using the four primary validation-selected EMA checkpoints without any weight update, we observe the neural 66-channel pose head before carrier constraints. All 256 token values still enter the decoder. We evaluate the same 1,400 eligible official validation clips, with the standard 40–192-frame centered crops and zero skips. Facing-local body MPJPE averages Euclidean errors over the 21 non-pelvis joints before global root translation and heading are applied. It therefore measures neural articulated reconstruction, not global-path accuracy or the usual root-relative metric, which retains heading error. In these four primary models, analytical recovery does not change the local body coordinates; this equality is checked directly.

As a separate diagnostic, we recover global joints after giving every model the same ground-truth root translation, heading, and height. These common-trajectory scores are oracle controls for comparing body outputs, not deployable reconstruction results. The diagnostic uses each checkpoint's exact training-matched runtime and normalization; it neither removes carrier values from decoder input nor selects a new model.

## 4.4 Matched alternative-carrier reconstruction study

We train five additional 8 × autoencoders from scratch: a fresh 224-learned/32-explicit primary-design baseline, an exact-height replacement with the same split, and torso, stance-anchor, and effector variants with 208 learned plus 48 explicit values. All retain the 24 root/yaw values and a total width of 256. The latter three retain the original carrier and add 16 motion-derived values; exact-height instead replaces the original height/support summaries with eight framewise heights. Extraction and correction rules are given in Appendix A.3.

These five runs share the official data, training-only normalization, loss weights, optimization schedule, seed 42, and 300-epoch duration. Shared initialization is copied with carrier-column meanings preserved when the learned width changes. Each EMA checkpoint is selected solely by validation global MPJPE, then evaluated on the same 4,210 eligible official test clips with zero skips. Because this is a separate matched training study, percentage changes use its fresh baseline, not the independent 8.936-mm primary result. No new text-to-motion or AIST++ runs are performed for these variants; comparisons concern complete allocation-and-recovery designs rather than an isolated signal or carrier-size effect.

## 4.5 Released reconstruction baselines and common coordinates

MotionGPT uses its released HumanML3D VQ-VAE at 4 × temporal downsampling [7]. MLD-1 uses its released HumanML3D VAE with one global latent slot [4]. A 192-frame crop produces 24 PIAE tokens, 48 MotionGPT discrete positions, and one MLD-1 global continuous slot. We retain each model's native representation and token budget, then convert reconstructed joints to metres in a common canonical HumanML3D frame.

A target-derived coordinate transform is shared by target and prediction. No prediction-specific root, heading, rigid, or Procrustes alignment is applied. Independent target recovery must agree within $5\times 10^{-6}$ m mean and $2 \times 10^{-5}$ m maximum error. This numerical check validates coordinate conversion and never adjusts predictions.

## 4.6 Reconstruction metrics and aggregation

Global MPJPE is the mean Euclidean joint-position error. Root-relative MPJPE subtracts each motion's own root translation; it does not remove heading error. Velocity and acceleration errors compare finite differences at 20 FPS. Contact-foot speed measures reconstructed foot movement during target contact. Penetration is the fraction of evaluated joints more than 5 mm below the ground plane. The terminal root error is the Euclidean distance between reconstructed and target root positions in all three coordinates at the final frame; it includes height error even when the planar path is exact.

Micro aggregation pools valid elements, so longer crops contribute more joint or temporal samples and contact speed is weighted by contact samples. Macro aggregation first computes a metric per clip and then weights clips equally. Lower values are better for every reconstruction metric.

## 4.7 Matched downstream text-to-motion protocol

We freeze each of the four autoencoders and train a matched hierarchical-flow generator based on Motion-HiFlow's temporal architecture [13]. All generators use a frozen CLIP ViT-B/32 text encoder, 256-channel input and output, a 384-wide nine-block transformer, effective batch size 64, the same initialization and pretrained backbone, and a 400k-step schedule. Every latent channel is predicted from text and noise, including carrier channels; no ground-truth motion or carrier is available at generation time. Latent normalization is fitted separately for each frozen autoencoder using training data only.

Validation FID selects among predeclared checkpoints. Final HumanML3D test evaluation uses the official MotionHiFlow evaluator for 20 replications and 4,224 evaluated records per replication. We report FID, R-precision, matching score, diversity, and multimodality. The four generators differ only in the frozen motion representation they must model. This is a downstream-task test with generators trained on HumanML3D, not zero-shot text-to-motion transfer to a new dataset. At each fixed temporal rate, carrier and no-carrier inputs have identical token counts and width.

### 4.8 Cross-dataset transfer protocol

To test whether the reconstruction effect transfers beyond HumanML3D, we evaluate the four frozen autoencoders on the official AIST++ v1.0 pose-test motions [14]. We use the 137 released fitted motions in the official pose-test intersection, recover the documented SMPL translation, standardize to the HumanML3D 22-joint skeleton, and subsample exactly from 60 to 20 FPS. HumanML3D training normalization remains frozen: no AIST++ sample is used for training, fine-tuning, checkpoint selection, or normalization fitting. The reconstruction crop and metrics are otherwise unchanged.

### 4.9 Optimization

Autoencoder training uses 300 epochs, batch size 128, AdamW with learning rate $10^{-4}$ and weight decay $10^{-4}$, five warm-up epochs, gradient clipping at 1.0, EMA decay 0.999, bfloat16 precision, and seed 42. Training-only normalization uses 10,000 sampled training windows. The architecture and complete loss configuration are given in Appendix A.

### 4.10 Efficiency measurement

We benchmark all four completed EMA autoencoders serially in one inference-only job on an NVIDIA A100-SXM4-80GB. The same official validation motions and FP32 runtime are used for each; TF32 is disabled. After 10 warm-up passes, we report the median of 50 CUDA-synchronized repetitions at 192 frames and batch size one. The timed path starts from normalized native features and includes the encoder, decoder, global-joint recovery, and contact probabilities. It excludes disk I/O, source-to-native conversion, normalization, device transfers, and checkpoint loading. Latent payload assumes four bytes per FP32 scalar and excludes masks and metadata. Parameter counts include the autoencoder only.

For text-to-motion, we aggregate the per-batch generation traces saved by the same 20-replication official-test evaluations used in Table 6. All variants use an A100-SXM4-80GB, batch size 32, 20 flow steps, the same guidance scale and hierarchy, and 7,008 generated motions per replication, including multimodality repeats. Generation time includes text encoding, flow sampling, autoencoder decoding, and native joint recovery, but excludes CPU conversion to the evaluator's 263-dimensional representation and evaluator scoring. Peak memory is PyTorch allocated GPU memory, not total process VRAM. We report Slurm training-job wall time separately as resource context; it includes setup, validation, and checkpoint I/O and is not a controlled kernel-throughput measure.

## 5 Results

The experiments connect four findings: selective preservation improves reconstruction at a fixed budget; the improvement reaches the neural body before correction; different explicit references favor different reconstruction outcomes; and the compressed representation supports generation and frozen transfer. The primary and targeted controls share the official test cohort. The body diagnostic uses validation clips, and the alternative-carrier study uses its own fresh matched baseline.

### 5.1 Selective preservation improves fixed-budget reconstruction

We first test whether explicit information earns its share of the token budget. In Table 2, the 8× carrier and no-carrier models both store 256 values per eight frames. The carrier reduces global MPJPE from 49.609 to 8.936 mm (82.0%) and root-relative error by 77.1%, with lower velocity, acceleration, contact-foot speed, and penetration errors. Terminal 3D root-position error falls from 48.135 mm to $9.24 \times 10^{-5}$ mm. The design therefore improves articulated and temporal fidelity alongside its structural trajectory preservation.

**Table 2:** Controlled HumanML3D micro aggregate over all 4,210 eligible official test clips. Each token has width 256; 1× uses eight times the latent storage of 8×. Lower is better. C denotes the physical carrier.

| Variant | Rate | Global (mm) | Root-rel. (mm) | Velocity (mm/s) | Accel. ($m/s^2$) | Contact (mm/s) | Penetration (%) | End root (mm) |
|---|---|---|---|---|---|---|---|---|
| 1× + C | 1× | **5.641** | **5.641** | **20.933** | **0.459** | **16.177** | **0.0035** | **0.00009** |
| 1× no C | 1× | 45.961 | 38.275 | 125.904 | 1.774 | 33.705 | 0.1091 | 37.863 |
| 8× + C | 8× | 8.936 | 8.884 | 62.678 | 1.368 | 24.608 | 0.0178 | **0.00009** |
| 8× no C | 8× | 49.609 | 38.846 | 165.057 | 2.089 | 42.332 | 0.1080 | 48.135 |

The benefit also appears without temporal compression: at 1 ×, adding the carrier reduces global MPJPE from 45.961 to 5.641 mm. Moving from 1 × to 8 × with the carrier then uses 87.5% fewer latent scalars at a 3.295-mm increase in global error. Thus, under the matched backbone and training recipe, increasing the number of fully learned temporal tokens does not substitute for the explicit structure: the compact carrier model reconstructs more accurately than the larger 1× learned-only reference.

## 5.2 Trajectory-aware controls clarify the allocation benefit

Table 3 compares PIAE with two strong trajectory-aware representations. The separately trained root/yaw-only control reaches 10.765 mm, recovering approximately 95.5% of the absolute no-carrier-to-full-carrier error reduction. This ratio describes the complete trained root-aware design, including its body prediction. The full 32-dimensional carrier reaches 8.936 mm, a further 17.0% improvement, showing the value of the combined height/support allocation and recovery beyond root/yaw alone. The root/yaw-only control has an exact planar path but retains 3.494 mm terminal *3D* root-position error because height remains learned. Its lower penetration also identifies a trade-off between the two designs.

The adapted ARDY-style explicit-global-root, root-conditioned-body control obtains 9.717 mm global MPJPE. Full PIAE reaches 8.936 mm, a 0.782-mm (8.0%) reduction, with the same 256-value token budget and common training/evaluation contract. Its root-relative error is also lower (8.884 versus 9.717 mm), as is velocity error (62.678 versus 67.702 mm/s). Both designs preserve the global endpoint to floating-point precision. PIAE's advantage therefore concerns articulated and temporal reconstruction beyond endpoint preservation, while using 32 rather than 40 explicit values per token.

**Table 3:** Targeted 8 × autoencoder controls on the same 4,210 HumanML3D test clips, with one 256-dimensional token per eight frames. Global, root-relative, velocity, contact speed, and penetration are micro aggregates; terminal 3D root-position error is the equal-clip mean. The ARDY-style row is the adapted root–body control described in Section 4.2. Lower is better.

| Variant | Learned /token | Explicit /token | Global (mm) | Root-rel. (mm) | Velocity (mm/s) | Contact (mm/s) | Penetration (%) | End root (mm) |
|---|---|---|---|---|---|---|---|---|
| 8× no carrier | 256 | 0 | 49.609 | 38.846 | 165.057 | 42.332 | 0.1080 | 48.135 |
| 8× root/yaw only | 232 | 24 | 10.765 | 10.507 | 78.569 | 29.365 | **0.0072** | 3.494 |
| 8× ARDY-style | 216 | 40 | 9.717 | 9.717 | 67.702 | 28.623 | 0.0395 | 0.00017 |
| 8× PIAE (full) | 224 | 32 | **8.936** | **8.884** | **62.678** | **24.608** | 0.0178 | **0.00009** |

## 5.3 The neural body improves before analytical correction

The neural-body diagnostic directly tests whether the gain extends beyond recovery of the retained trajectory. Table 4 observes the frozen primary checkpoints on the separate validation cohort in Section 4.3. At 8×, facing-local body MPJPE falls from 40.011 to 9.352 mm (76.6%); at 1×, it falls from 39.262 to 5.947 mm (84.9%). The carrier model improves on every paired clip in each comparison. These body coordinates are neural outputs and are not analytically overwritten in the primary model.

**Table 4:** Frozen-checkpoint diagnostic on 1,400 eligible HumanML3D *validation* clips, with zero skips. Body error measures the neural facing-local 21-joint output before correction, excluding the pelvis, global translation, and global heading. The final column recovers all 22 global joints with the same true root translation, heading, and height for every model; it is an oracle diagnostic, not a deployable score. Values are micro aggregates in mm. All 256 token values, including any carrier, remain available to the decoder.

| Variant | Learned | Explicit | Neural body | Common-trajectory global |
|---|---|---|---|---|
| 8× no carrier | 256 | 0 | 40.011 | 38.193 |
| 8× + carrier | 224 | 32 | 9.352 | 8.927 |
| 1× no carrier | 256 | 0 | 39.262 | 37.477 |
| 1× + carrier | 248 | 8 | 5.947 | 5.677 |

On these same validation clips, the learned-only model has 51.307 mm normal global MPJPE. Replacing its root translation, heading, and height with the true values after decoding reduces the error to 38.193 mm, versus 8.927 mm for the carrier model under the identical true trajectory. Post-hoc path repair therefore does not reproduce the trained carrier model's body quality. The key finding is better *carrier-conditioned neural body reconstruction*: the explicit/learned design provides a more accurate neural articulated output as well as a reliable global reference. All token values remain available to the decoder in this diagnostic; Appendix A specifies the observed heads and their training roles.

## 5.4 Different explicit references favor different outcomes

Table 5 asks whether selected non-root information can add value while preserving the total token width. In this independent matched study, the fresh primary-design baseline achieves 8.977 mm global MPJPE. Torso coordinates improve it to 8.504 mm (5.3%), stance anchors to 8.667 mm (3.5%), and effector endpoints to 8.750 mm (2.5%). The three variants replace 16 learned values with 16 additional carrier values; they do not enlarge the 256-dimensional token or change its 8× temporal rate.

**Table 5:** Separate matched alternative-carrier study on the same 4,210 eligible HumanML3D test clips; all models complete 300 epochs, use validation-selected EMA checkpoints, and have zero skips. Every row has 256 values per eight frames. Errors and physical metrics are micro aggregates; lower is better. $\Delta$is the percentage reduction in global MPJPE relative to this table's fresh baseline (positive is better), not the independent 8.936-mm primary checkpoint. Contact speed is measured during target contact.

| Carrier design | Learned /token | Explicit /token | Global (mm) | Root-rel. (mm) | Contact (mm/s) | Penetration (%) | $\Delta$ (%) |
|---|---|---|---|---|---|---|---|
| Primary design, fresh baseline | 224 | 32 | 8.977 | 8.924 | 24.872 | 0.01765 | – |
| Exact-height replacement | 224 | 32 | 10.133 | 10.133 | 29.282 | 0.04016 | −12.9 |
| Primary + torso XY | 208 | 48 | **8.504** | **8.441** | 25.628 | 0.01923 | +5.3 |
| Primary + stance anchors | 208 | 48 | 8.667 | 8.618 | **20.588** | **0.01110** | +3.5 |
| Primary + effector XY | 208 | 48 | 8.750 | 8.693 | 25.576 | 0.01690 | +2.5 |

Different references improve different outcomes. Torso coordinates give the best position accuracy, while stance anchors reduce contact-foot speed by 17.2% and penetration by 37.1% relative to the fresh baseline. All five designs retain numerically negligible terminal root error. Storing exact heights while replacing the original height/support summaries instead worsens MPJPE by 12.9%. This replacement also removes the learned height residual and contact prior. The comparison shows the value of evaluating information and its recovery together: retaining a quantity more precisely need not produce a better complete motion representation.

These results extend selective preservation beyond trajectory signals through bounded support references and torso coordinates. Both improve the allocation-and-recovery system while retaining the same total width; their different metric profiles make the choice dependent on the intended reconstruction outcome. Their corrections act on body joints, as detailed in Appendix A.3. The following T2M, efficiency, and AIST++ studies use the original 32-dimensional carrier, keeping its broader evaluation separate from this reconstruction-focused comparison.

### 5.5 Generated carrier tokens improve matched text-to-motion

An explicit carrier is useful for generation only if its values can be modeled jointly with the rest of the token. In Table 6, every channel is predicted from text and noise; no input motion supplies a true carrier. With the four original AEs frozen, matched 8× generators improve FID from 0.506 to 0.163 (67.7%) when the carrier is used, alongside better R-precision and matching score. At this fixed temporal rate, both representations have the same number of 256-dimensional tokens, so the FID difference cannot be explained by a shorter sequence. The compressed carrier also has 11.1% lower FID than 1× without a carrier. Published MotionGPT and MLD rows provide context, not reruns of those systems under our generator.

**Table 6:** Matched HumanML3D text-to-motion evaluation over 20 official-evaluator replications. FID and matching score are lower; R-precision is higher. Real-motion statistics are evaluator references, not a generated model. MotionGPT and MLD rows are published HumanML3D reference values and are not rerun under the adapted PIAE interface.

| Representation | FID↓ | R@1↑ | R@2↑ | R@3↑ | Match↓ | Diversity | Multimod. |
|---|---|---|---|---|---|---|---|
| 1× + carrier | 0.282 | 0.478 | 0.668 | 0.770 | 3.148 | 9.036 | **1.343** |
| 1× no carrier | 0.183 | 0.507 | 0.702 | 0.800 | 2.972 | **9.247** | 0.921 |
| 8× + carrier | **0.163** | **0.521** | **0.712** | **0.808** | **2.887** | 9.206 | 1.133 |
| 8× no carrier | 0.506 | 0.503 | 0.698 | 0.798 | 3.045 | 9.150 | 1.060 |
| Real motions | – | 0.511 | 0.702 | 0.796 | 2.967 | 9.467 | – |
| *Published HumanML3D reference values* | | | | | | | |
| MotionGPT [7] | 0.232 | 0.492 | 0.681 | 0.778 | 3.096 | 9.528 | 2.008 |
| MLD [4] | 0.473 | 0.481 | 0.673 | 0.772 | 3.196 | 9.724 | 2.413 |

Reconstruction accuracy and generative usability are therefore related but distinct. The 1× carrier model is the best reconstructor, yet its downstream FID is 53.9% higher than the 1× no-carrier model. At each fixed rate, adding the carrier changes token contents, not sequence length or total width. The 8× result consequently demonstrates a benefit from the structured interface at the intended compressed rate. The contrasting 1× outcome makes downstream evaluation informative in its own right: reconstruction quality alone does not predict how easily the complete token distribution can be generated. Possible explanations, including redundancy and carrier–pose agreement, are considered in the discussion.

### 5.6 A shorter temporal interface delivers measured efficiency

The temporal-rate comparison also tests the practical reason to compress motion before generation. Table 7 reports actual FP32 latent payload and matched inference measurements. For a 192-frame clip, the 8× representation contains 24 tokens and 6,144 scalars (24 KiB), whereas the 1× representation contains 192 tokens and 49,152 scalars (192 KiB). This is an eightfold latent-storage reduction at fixed token width, not an eightfold reduction relative to the 70-channel input; the latter is 2.1875-fold. With a full-attention layer on these latent positions, the potential motion-token interaction count falls by 64×. That count is a property of the attention matrix, not a measured whole-model speedup.

**Table 7:** Measured representation and inference efficiency on an A100-SXM4-80GB. Latent payload is FP32 for a 192-frame clip. AE latency is median full-path inference at batch size one. Text-to-motion time is amortized per generated motion at batch size 32; peak GPU memory is PyTorch allocated memory during generation. The timing scopes are specified in the protocol.

| Variant | Tokens /192 frames | Latent (KiB) | AE params (M) | AE time (ms) | T2M time (ms/motion) | T2M rate (motions/s) | T2M peak (GiB) |
|---|---|---|---|---|---|---|---|
| 1× + carrier | 192 | 192 | 22.598 | 82.75 | 17.564 | 56.94 | 0.887 |
| 1× no carrier | 192 | 192 | 22.602 | 49.55 | 17.195 | 58.16 | 0.887 |
| 8× + carrier | 24 | 24 | 22.586 | 81.10 | 12.674 | 78.90 | 0.703 |
| 8× no carrier | 24 | 24 | 22.602 | 48.34 | 12.673 | 78.91 | 0.703 |

Compared with 1× plus carrier, 8× plus carrier reduces measured generation time per motion by 27.8% and peak allocated GPU memory by 20.7%; throughput rises from 56.94 to 78.90 motions/s. The analogous no-carrier time reduction is 26.3%. Adding the carrier at 8× changes measured generator time negligibly (12.673 to 12.674 ms/motion) while improving FID by 67.7%. Carrier extraction and recovery add work to the autoencoder itself: the full 8× carrier AE takes 81.10 ms at batch size one, versus 48.34 ms without the carrier, and is 2.0% faster than the 1× carrier AE. The measured deployment gain is therefore concentrated where compression is intended to help: the downstream model operating on a shorter sequence.

The payload, attention-pair count, and measured runtime describe different aspects of that benefit: stored values, comparable-layer interactions, and complete generative inference. Appendix B.3 reports recorded training wall time separately as resource context for the completed experiments.

## 5.7 Reconstruction gains transfer to AIST++ without adaptation

Table 8 applies the original frozen checkpoints to official AIST++ dance motions, with no fine-tuning or normalization refitting. At 8×, the carrier reduces global MPJPE from 189.594 to 24.544 mm (87.1%) and root-relative error by 72.5%. Contact-foot speed is also lower, while penetration is slightly higher (0.0187% versus 0.0150%). The reconstruction advantage therefore transfers across motion datasets, with the table retaining the distinct outcomes of each physical proxy.

**Table 8:** Zero-shot reconstruction on 137 official AIST++ v1.0 pose-test motions. Frozen HumanML3D checkpoints and normalization are used without adaptation. Lower is better.

| Variant | Global (mm) | Root-rel. (mm) | Velocity (mm/s) | Accel. (m/s$^2$) | Contact (mm/s) | Penetration (%) | End root (mm) |
|---|---|---|---|---|---|---|---|
| 1× + carrier | **10.588** | **10.588** | **94.506** | **2.379** | **23.218** | **0.0004** | **0.00003** |
| 1× no carrier | 173.037 | 77.057 | 540.660 | 8.902 | 31.685 | 0.0022 | 211.151 |
| 8× + carrier | 24.544 | 24.249 | 310.285 | 7.378 | 28.172 | 0.0187 | **0.00003** |
| 8× no carrier | 189.594 | 88.283 | 696.136 | 9.967 | 36.129 | 0.0150 | 191.215 |

The advantage survives a substantial change in motion domain while preserving the original model and statistics. Together with the in-domain neural-body diagnostic and the generated-carrier experiment, this provides a separate test of practical usefulness: the reconstruction interface can be reused on a new motion dataset without adaptation.

## 5.8 Comparison with released and published autoencoders

We also evaluate released MotionGPT and MLD-1 autoencoders under the shared crop and coordinate contract (Table 9). PIAE has lower global and root-relative MPJPE, velocity error, and contact-foot speed in these reruns. The comparison supplies external reconstruction context, but the models retain different native token rates and budgets; the equal-budget PIAE controls above address the allocation question more directly.

**Table 9:** HumanML3D micro aggregate over all 4,210 eligible test clips under the shared reconstruction protocol. Lower is better.

| Model | Global (mm) | Root-relative (mm) | Velocity (mm/s) | Accel. (m/s$^2$) | Contact speed (mm/s) | Penetration (%) | End root (mm) |
|---|---|---|---|---|---|---|---|
| Ours 8× | **8.936** | **8.884** | **62.678** | **1.368** | **24.608** | **0.018** | **0.000** |
| MotionGPT VQ 4× | 172.085 | 55.987 | 212.885 | 2.934 | 105.854 | 0.345 | 251.731 |
| MLD-1 | 104.408 | 43.924 | 223.479 | 2.271 | 66.276 | 0.038 | 126.876 |

Appendix B retains the equal-clip aggregates, per-clip distributions, and published reconstruction context. Literature values with different skeletons and protocols are not treated as a controlled SOTA ranking.

## 5.9 Qualitative reconstruction

Figure 2 shows eight synchronized poses from a 192-frame example for ground truth, PIAE, MotionGPT, and MLD-1, using the same coordinates without prediction-specific alignment. It illustrates body reconstruction and global placement, rather than replacing the aggregate comparisons. The per-clip error distribution is retained in Appendix B.

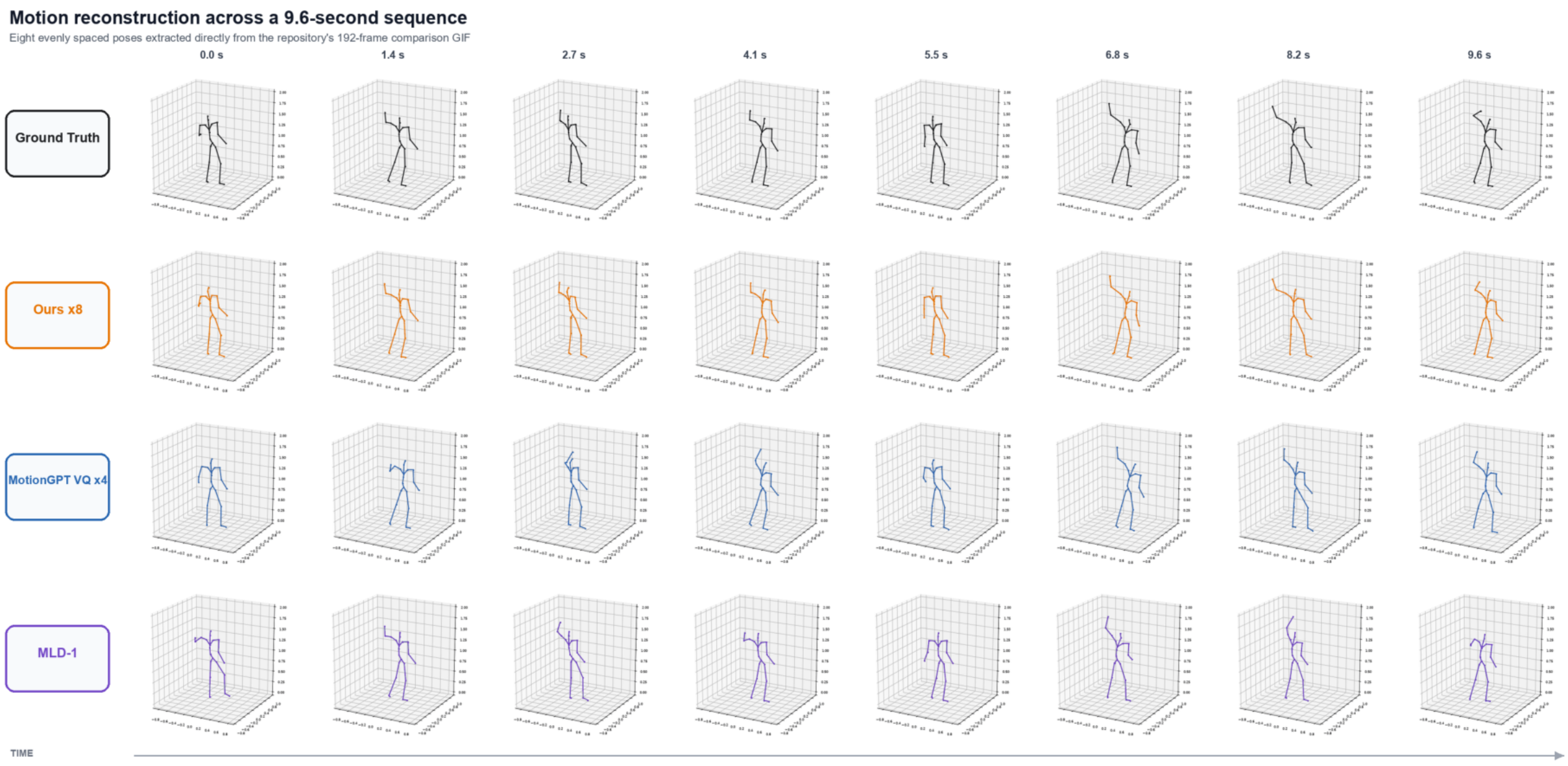

**Figure 2:** Synchronized qualitative reconstruction for ground truth, Ours, MotionGPT, and MLD-1, with no prediction-specific alignment.

# 6 Discussion

**The value lies in how the budget is allocated.** The same 256-value token can support substantially different reconstruction and generation quality depending on what its values represent. PIAE reserves a small fraction of that budget for reliable kinematic references and uses the remaining capacity for learned features. The improvement therefore reflects the organization of the representation and its reconstruction paths, rather than an increase in stored information. The $1\times$ learned-only model's larger temporal payload also does not recover the $8\times$ carrier model's reconstruction quality under the matched recipe. These results identify selective preservation as an effective alternative to assigning all available dimensions to an unconstrained learned code.

**A reliable reference accompanies a better neural body prediction.** Trajectory recovery and neural reconstruction play complementary roles. Root/yaw-only preservation already produces a strong reconstruction baseline, and the full carrier further improves articulated accuracy. Crucially, the frozen body-path diagnostic observes the neural prediction itself: body error is 76.6% lower at $\times 8$ before correction, and giving both models the same true translation, heading, and height retains the advantage. Thus the full model does more than place a similar-quality body on the right path; it produces a more accurate articulated body. A plausible explanation is that accurate reference information both conditions the decoder and prevents trajectory mistakes from contaminating post-recovery global reconstruction errors during training. This explains the design intuition behind learning complementary detail around reliable explicit references; the diagnostic establishes the body-quality result, while the mechanisms remain a subject for further study.

**What to preserve depends on what the reference supports.** Root/yaw velocities are important because their integration controls global placement over time. Torso and stance information serve a different role: they provide shared spatial references for connected upper-body joints or supported feet. Their different outcomes are informative. Torso produces the lowest positional error in the follow-up study, while stance anchors give the best contact-foot speed and penetration. The exact-height variant replaces summarized height, support information, and learned within-block height detail with framewise height storage, and performs less accurately overall. The resulting design lesson is to choose preserved information together with its neural conditioning, recovery rule, and displaced learned capacity. Greater explicit precision on one quantity is not a substitute for a useful division of roles across the complete motion.

**A generative interface must organize predictable information.** In reconstruction, explicit values are extracted from the input. In text-to-motion, the generator must instead learn their joint distribution with the body features. The 8 × FID improvement therefore demonstrates generative usability of the complete structured token, with identical token counts and widths in the carrier/no-carrier comparison. The different ranking at 1× further shows that the best reconstruction interface need not be the best generative interface. Redundancy, carrier–pose agreement, and generator optimization are candidate explanations for this rate-dependent behavior. The empirical lesson is to assess the representation at its intended temporal rate and downstream objective, rather than infer generation quality from reconstruction alone.

**Transfer and efficiency make the representation practically useful.** The AIST++ evaluation applies the original checkpoints and HumanML3D statistics unchanged to dance motions, demonstrating frozen reconstruction transfer to another motion dataset. Separately, the shorter 8 × carrier sequence improves generation throughput and peak memory relative to its 1× counterpart. These are complementary advantages: transfer tests whether the representation remains useful on new motions, while efficiency measures the practical benefit of reducing the generator's temporal interface. Table 10 summarizes the evidence supporting these conclusions.

**Table 10:** Complementary evidence for the fixed-budget representation. These are the existing studies; downstream-task use and cross-dataset reconstruction address different questions.

| Design question | Matched evidence | Main finding |
|---|---|---|
| Neural reconstruction | Frozen primary carrier/no-carrier checkpoints; body outputs before correction and common true trajectory | Better articulated output persists beyond trajectory restoration. |
| Preserved information | Root/yaw-only, primary carrier, adapted root hybrid, and separate alternative-carrier study | Signal allocation and recovery matter; torso and stance favor different outcomes. |
| Generative use | Frozen original AEs and matched HumanML3D T2M generators predicting all token channels | The 8× carrier improves FID; reconstruction and generation rank the interfaces differently. |
| Motion dataset | Frozen HumanML3D AEs on official AIST++ poses with unchanged normalization | Reconstruction gains transfer without adapting the autoencoder or its statistics. |
| Temporal efficiency | Same-width 1× and 8× interfaces, measured generation time and peak memory | Shorter carrier sequences improve downstream time and memory while retaining accurate motion reconstruction. |

## 6.1 Limitations and future work

The experiments establish the effect of the complete carrier-conditioned representation under a matched backbone and training recipe. Carrier signals and their allocations are manually specified; the study does not establish an optimal or automatic selection rule. The body diagnostic retains the carrier as decoder input; it does not isolate the learned 224-dimensional subcode or separate conditioning from changes in the training-loss pathway. Alternative-carrier corrections also act on body joints, so that study compares complete allocation-and-recovery designs. Its additional variants have not been evaluated in T2M or AIST++, and the AIST++ study has no separate pre-correction body diagnostic. These distinctions delimit which component-level explanations follow from each experiment.

The reconstruction studies use one training seed and crops capped at 192 frames; evaluator replications quantify sampling variation rather than independent training variation. The learned-only control is a matched carrier-removal ablation, not an independently optimized family of fully learned AEs. Its 1 × result does not identify the cause of that model's remaining reconstruction error. The ARDY-style control evaluates an adapted autoencoder, while the published reference rows retain their original protocols. Continuous scalar counts specify equal-precision payload rather than equal entropy-coded bitrate with discrete tokenizers. PIAE is kinematic: exact recovery of input-derived trajectory values does not guarantee physically consistent generated motion, and lower FID is not a dynamics test. The generation study uses one matched generator family on HumanML3D; AIST++ tests reconstruction transfer, not cross-dataset generation.

Future work can investigate independent training-seed variation, separate the effects of neural conditioning and structured recovery, and extend predicted-carrier consistency to physical generation objectives. The fixed-budget formulation also motivates task-dependent signal and capacity selection, and applications in other temporal domains with explicit motion or reference information. The present work provides the implemented motion design and experimental foundation for those extensions.

# 7 Conclusion

A compact representation can benefit from deciding what to preserve explicitly and what to learn. PIAE implements this division within a single fixed-width motion token: selected kinematics provide reliable references, learned features encode complementary detail, and both inform neural decoding alongside structured recovery. At $8\times$ temporal compression, this design improves global reconstruction and neural body prediction within the same 256-value budget. The body-quality advantage persists before analytical correction and when both models receive the same true trajectory, establishing a benefit beyond path replacement alone.

The representation also supports use beyond encoding its training motions. Matched text-to-motion generation improves when the entire compressed token, including its carrier, must be predicted; the original frozen AE transfers its reconstruction advantage to AIST++ without adaptation. Alternative carriers show that positional accuracy and support quality favor different explicit references, while the rate-dependent generation results distinguish reconstruction fidelity from generative utility. Measured downstream time and memory savings demonstrate the practical value of the shorter temporal interface.

The resulting takeaway is to allocate limited representation capacity according to the role information plays in reconstruction and downstream modeling, rather than assume that every value must be learned through the bottleneck. PIAE demonstrates this approach through a concrete carrier design, controlled motion experiments, and complementary evidence from neural-body reconstruction, generated tokens, and frozen cross-dataset transfer.

# Acknowledgements

We are grateful to the KAUST Academy for its generous support, and especially to Prof. Sultan Albarakati, whose support made this work possible.

# A Implementation Details

## A.1 Carrier normalization and constrained decoding

The 16 planar-velocity values are divided by 2, the eight angular-velocity values by 4, and the four height statistics by 2. The four support probabilities are unchanged. These fixed scales are inverted before analytical expansion; they do not quantize the values. Native motion channels separately use training-set means and standard deviations.

For an eight-frame block, height expansion starts with linear interpolation between the stored endpoints. An equal correction is added to the six interior values so that the resulting block mean equals the stored mean. In normalized native coordinates, the decoder adds $0.25\,\Pi(u)$, where $u$ is its predicted height residual and $\Pi$ sets both endpoints to zero and subtracts the interior mean. This preserves endpoints and mean while allowing learned interior variation. The stored height standard deviation has no separate hard constraint.

For foot $j$, soft input support is $\sigma((0.04 - P_{\mathrm{t,j,z}})/0.01)\,\sigma((0.12 - s_{\mathrm{t,j}})/0.03)$, where $s_{\mathrm{t,j}}$ is foot speed in m/s and heights are in metres. Carrier expansion repeats the four block-average support values over the block and clips them to $[0.02, 0.98]$. Their logits are added to the predicted contact logits with weight 1.0. Contact output remains learned; this operation is not an inverse reconstruction of the original per-frame contacts.

## A.2 Complete training objective

**Table 11:** Loss configuration. Squared-error terms average over valid scalar entries; contact-gated terms normalize by their valid contact weights.

| Term | Weight | Definition |
|---|---|---|
| Local pose | 2 | Squared native local-joint coordinate error |
| Global pose | 10 | Squared reconstructed global-joint coordinate error |
| Planar-root velocity | 0.5 | Squared planar-root velocity error |
| Yaw velocity | 0.1 | Squared yaw angular-velocity error |
| Root height | 2 | Squared root-height error |
| Contact classification | 0.05 | Binary cross-entropy on four contact outputs |
| Joint velocity | 0.05 | Squared first-difference error, scaled by $f$ |
| Joint acceleration | 0.0005 | Squared second-difference error, scaled by $f^2$ |
| Bone length | 2 | Mean absolute predicted–target bone-length difference |
| Foot sliding | 0.1 | Squared planar foot speed during target contact |
| Contact anchoring | 10 | Squared foot-position error during target contact |
| Penetration | 5 | Mean squared negative joint height |
| Terminal SE(2) | 1 | Duration-normalized endpoint position and yaw error |

The objective is the weighted sum of Table 11. Joint velocities and accelerations match target derivatives. Foot sliding is a separate magnitude penalty applied only where consecutive frames both indicate target contact. Contact anchoring additionally matches target foot positions. The terminal term squares planar endpoint error divided by clip duration and adds squared wrapped yaw error divided by duration. Ground penetration is penalized by $\min(P_z, 0)^2$ during training; the evaluation metric instead counts joints below $-5$ mm.

## A.3 Alternative-carrier extraction and recovery

The follow-up study leaves the primary model unchanged and implements four alternatives. All signals are computed from the same unnormalized native input; no external observations or target contact annotations enter the carrier. All variants concatenate explicit and learned values before neural decoding and retain framewise root/yaw velocities.

**Exact-height replacement (224 learned + 32 explicit).** The final eight explicit values store the eight framewise root heights instead of the original four height statistics and four support summaries. Heights are scaled by 1 /2 for storage and recovered exactly. The learned height residual and additive contact-logit prior are disabled. This is therefore a replacement of both information and recovery rules, not a height-only ablation.

**Torso XY (208 learned + 48 explicit).** Sixteen extra values store the facing-local planar coordinates of the upper-trunk joint (SPINE3) at all eight frames. The difference between that anchor and the predicted SPINE3 position is applied with weights 1/3, 2/3, and 1 to SPINE1, SPINE2, and SPINE3, respectively. The same unit-weight correction is applied to neck, head, collars, shoulders, elbows, and wrists. Vertical coordinates are unchanged. Thus a torso anchor affects a set of connected upper-body coordinates rather than only replacing one joint.

**Stance anchors (208 learned + 48 explicit).** For each foot and each four-frame half-block, the carrier stores the XYZ position at the frame of maximum motion-derived soft support, together with that confidence. Two feet, two anchors, and four values per anchor require 16 dimensions. Anchor XY is expressed in the block's initial root frame; Z remains a ground-referenced height. Anchors with confidence below 0.1 are set to zero. At decoding, the appropriate half-block anchor is transformed back to the current facing-local frame. Its displacement from the predicted foot is norm-clipped to 0.25 m, then multiplied by 0.5 times stored confidence times predicted foot-contact probability. This bounded correction is added to the foot and half of it to the corresponding ankle. The rule provides a soft support reference, not a hard foot lock or a guarantee for generated carrier values.

**Effector XY (208 learned + 48 explicit).** Sixteen extra values store the first- and last-frame facing-local XY coordinates of both wrists and both feet. The decoder adds the linearly interpolated endpoint discrepancy to those four joints over the eight-frame block. This preserves the stored planar endpoints while leaving interior detail and vertical motion learned.

## A.4 Diagnostic scope and checkpoint provenance

The neural-body diagnostic uses only the original four frozen checkpoints and their training-matched source snapshots. Observing the raw heads leaves normal inference unchanged; checks confirm identical local body coordinates before and after the primary carrier constraints. Removing all constraints is a different question: raw global MPJPE is 1,236.495 mm for the 8× carrier checkpoint and 1,085.557 mm for the 1× carrier checkpoint, because their overwritten root/yaw heads are not independently supervised trajectory outputs. At 8×, the raw height head is also a constrained residual rather than an absolute height predictor. These raw-global diagnostic values do not describe the deployed model and must not be confused with the facing-local body scores in Table 4.

The alternative-carrier validation selections are epochs 299 (fresh baseline), 298 (exact height), 296 (torso), 300 (stance), and 299 (effector). Training completion at epoch 300 is verified for every run. Source, normalization, split, selected-checkpoint hashes, and aggregate scores are retained in the accompanying metadata. No diagnostic result is used to select a new checkpoint, and no downstream result is claimed for the alternative carriers.

## A.5 Result provenance and reproducibility

The source package archives aggregate JSON for four-way reconstruction (4,210 official HumanML3D clips, zero skips, and the crop and coordinate contract), the two targeted 8 × controls (the same test cohort and validation-selected EMA checkpoint identities), matched text-to-motion (selected checkpoints and 20 evaluator replications), AIST++ transfer (137 official pose-test motions, asset checksums, and zero skips), and efficiency (latent counts, parameters, timings, memory, Slurm wall time, and audit SHA-256). Follow-up metadata additionally cover the four-model frozen validation diagnostic and the five-model matched alternative-carrier study, with contracts, completion records, validation selections, and test JSONs. These datasets and cohorts remain distinct in the reported tables. The MotionGPT and MLD baselines use their released HumanML3D VQ-VAE and FID041 VAE checkpoints, respectively. Efficiency and decoder-path audits are inference-only. SI-to-mm conversions use stored metrics. Model and evaluation code are provided; HumanML3D and AIST++ remain under their own data terms.

# B Additional Reconstruction and Resource Context

The main results follow the fixed-budget allocation, body-path, and generative-interface questions. This appendix retains the broader benchmark and resource evidence without conflating it with the matched representation comparisons.

## B.1 Published reconstruction context and aggregation

Table 12 provides the broader published MPJPE context. Values follow their respective papers' protocols and representations; PRISM v4, for example, evaluates a shared motion135/SMPL-H forward-kinematics skeleton, whereas our shared comparison uses canonical HumanML3D joints. These values supply context, not a common-protocol performance ranking.

**Table 12:** Published HumanML3D reconstruction MPJPE at 4 × and 8× temporal compression. Lower is better; protocols follow the respective papers.

| Model | Temporal compression | MPJPE (mm) |
|---|---|---|
| Ours | 8× | **8.936** |
| PRISM [11] | 4× | 9.72 |
| MoMask + ParTY temporal modules [10] | 4× | 11.0 |
| MoMask + ParTY temporal modules [10] | 8× | 14.0 |
| ParTY holistic temporal-aware VQ-VAE [10] | 8× | 19.0 |
| MoMask RVQ-VAE [8] | 4× | 29.5 |
| MoMask VQ-VAE, larger window [10] | 8× | 55.0 |
| T2M-GPT VQ-VAE [6, 8] | 4× | 58.0 |

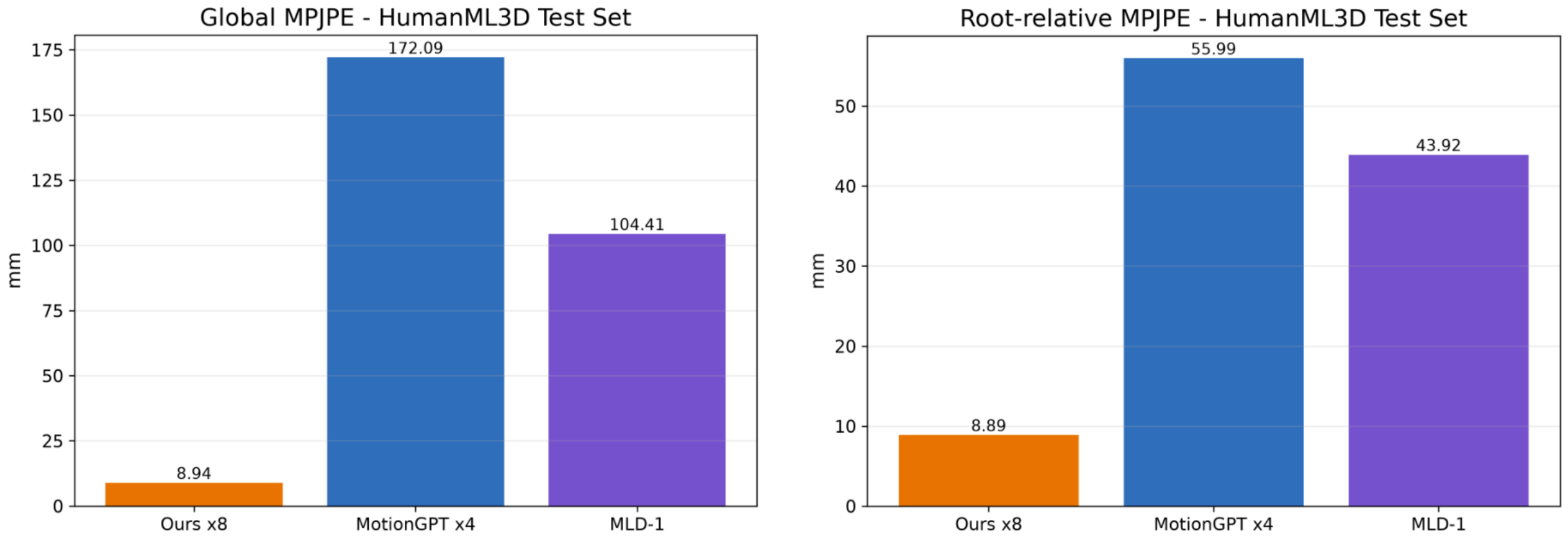

**Figure 3:** Global and root-relative MPJPE on the eligible official HumanML3D test clips under the shared protocol.

The equal-clip macro aggregate follows the same ranking (Table 13), indicating that the main result is not produced only by longer clips.

**Table 13:** HumanML3D macro aggregate over the same 4,210 clips, weighting every clip equally. Lower is better.

| Model | Global (mm) | Root-relative (mm) | Velocity (mm/s) | Accel. ($m/s^2$) | Contact speed (mm/s) | Penetration (%) | End root (mm) |
|---|---|---|---|---|---|---|---|
| Ours 8× | **8.995** | **8.933** | **66.535** | **1.500** | **31.624** | **0.022** | **0.000** |
| MotionGPT VQ 4× | 153.652 | 52.816 | 211.865 | 3.044 | 117.709 | 0.280 | 251.731 |
| MLD-1 | 90.506 | 39.362 | 212.416 | 2.369 | 84.242 | 0.036 | 126.876 |

## B.2 Per-clip reconstruction distribution

Figure 4 reports per-clip global MPJPE for the released-model comparison on the eligible HumanML3D test cohort. It complements the micro and macro aggregate tables; it does not substitute for independent training-seed uncertainty.

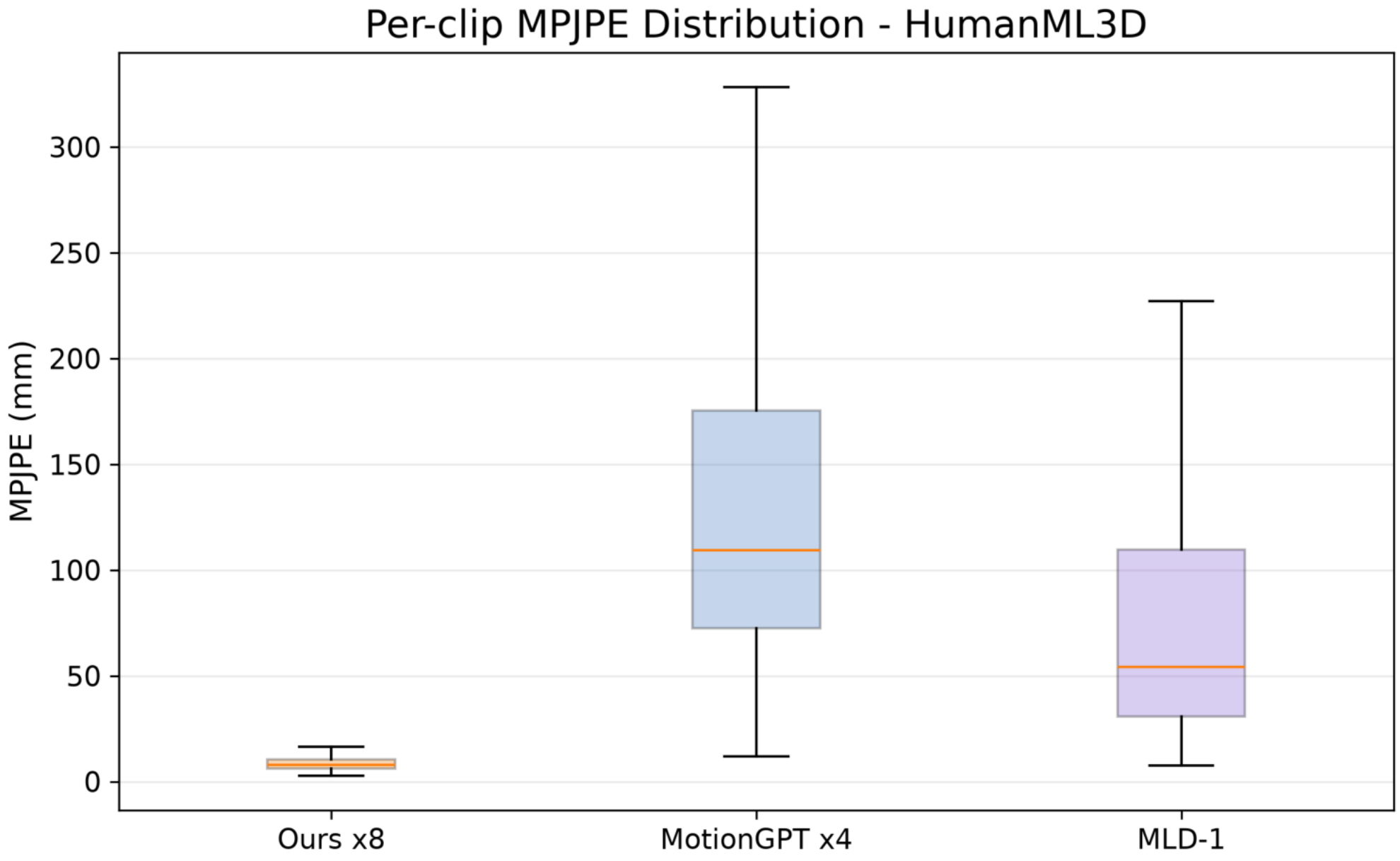

**Figure 4:** Distribution of per-clip global MPJPE across the eligible HumanML3D test set.

## B.3 Recorded training wall time

**Table 14:** Recorded A100 job wall time for the completed runs. AE jobs trained for 300 epochs; T2M totals include the 400k-step main run and its continuation. Values include setup, validation, and checkpoint I/O and are not isolated model-step timings.

| Variant | AE job (GPU-h) | T2M jobs (GPU-h) |
|---|---|---|
| 1× + carrier | 6.73 | 31.22 |
| 1× no carrier | 6.25 | 26.24 |
| 8× + carrier | 5.75 | 27.65 |
| 8× no carrier | 5.33 | 23.76 |

The wall-time comparison in Table14 provides resource context but should not be interpreted as a controlled training-throughput benchmark: runs were scheduled separately and the downstream jobs had different continuation durations. We do not infer the theoretical 64× attention-pair reduction from these timings; the measured generator effect is substantially smaller because text encoding, decoder operations, and other work remain.